\documentclass[runningheads]{llncs}
\usepackage[T1]{fontenc}
\usepackage{graphicx}
\usepackage{amssymb}
\usepackage{multirow}
\usepackage{cleveref}
\usepackage{booktabs}

\begin{document}
\title{Prompt2LVideos: Exploring Prompts for Understanding Long-Form Multimodal Videos}
%
%

\author{Soumya Shamarao Jahagirdar\inst{1}\orcidID{0000-0002-3460-9151} \and
Jayasree Saha\inst{1}\orcidID{0000-0002-0839-1849} \and
C V Jawahar\inst{1}\orcidID{0000-0001-6767-7057}}
\authorrunning{Jahagirdar et al.}
%
\institute{Center for Visual Information Technology, IIIT Hyderabad, India \\ 
\email{\{soumya.jahagirdar,jayashree.saha\}@research.iiit.ac.in} \\
\email{jawahar@iiit.ac.in}
}

%
\maketitle              
\begin{abstract}
Learning multimodal video understanding typically relies on datasets comprising video clips paired with manually annotated captions. However, this becomes even more challenging when dealing with long-form videos, lasting from minutes to hours, in educational and news domains due to the need for more annotators with subject expertise. Hence, there arises a need for automated solutions. Recent advancements in Large Language Models (LLMs) promise to capture concise and informative content that allows the comprehension of entire videos by leveraging Automatic Speech Recognition (ASR) and Optical Character  Recognition (OCR) technologies. ASR provides textual content from audio,  while OCR extracts textual content from specific frames. This paper introduces a dataset comprising long-form lectures and news videos. We present baseline approaches to understand their limitations on this dataset and advocate for exploring prompt engineering techniques to comprehend long-form multimodal video datasets comprehensively.
\keywords{Long-form videos  \and Prompt Engineering \and Large Language Models.}
\end{abstract}

\section{Introduction}
Learning representations for video-text fusion poses a significant challenge yet holds paramount importance for comprehensive video comprehension across various real-world applications such as video retrieval, captioning, and video based  document retrieval analysis. While multimodal contrastive learning leveraging extensive web-scale datasets has demonstrated efficacy in image-text representation tasks, its application in the video-language domain still needs to be explored, particularly in educational and news contexts. A primary hurdle hindering further exploration in this domain is the need for high-quality video-language datasets suitable for large-scale pretraining. Existing research predominantly relies on datasets such as HowTo100M \cite{howto100m_iccv_2019}, MSR-VTT\cite{msrvtt_cvpr_2016}, YOUCOOK2\cite{you_cook_2_aaai}, etc., utilizing automatically generated transcriptions via automatic speech recognition (ASR) or annotated captions. However, these datasets primarily consist of instructional videos focusing on action understanding. The distinct challenges presented by educational and news video content necessitate dedicated attention and dataset curation efforts to facilitate advancements in video-text representation learning within these domains.
In education, the focus typically diverges from action understanding, instead emphasizing the transmission of information through mediums like blackboards or PowerPoint presentations. Extracting meaningful insights from these sources poses challenges; solely relying on ASR is insufficient, and acquiring annotations can be complex due to the specialized subject matter knowledge required. Conversely, news videos dynamically convey information through live updates and on-screen graphics. 

\begin{figure*} 
    \centering
    \includegraphics[width=0.85\linewidth]{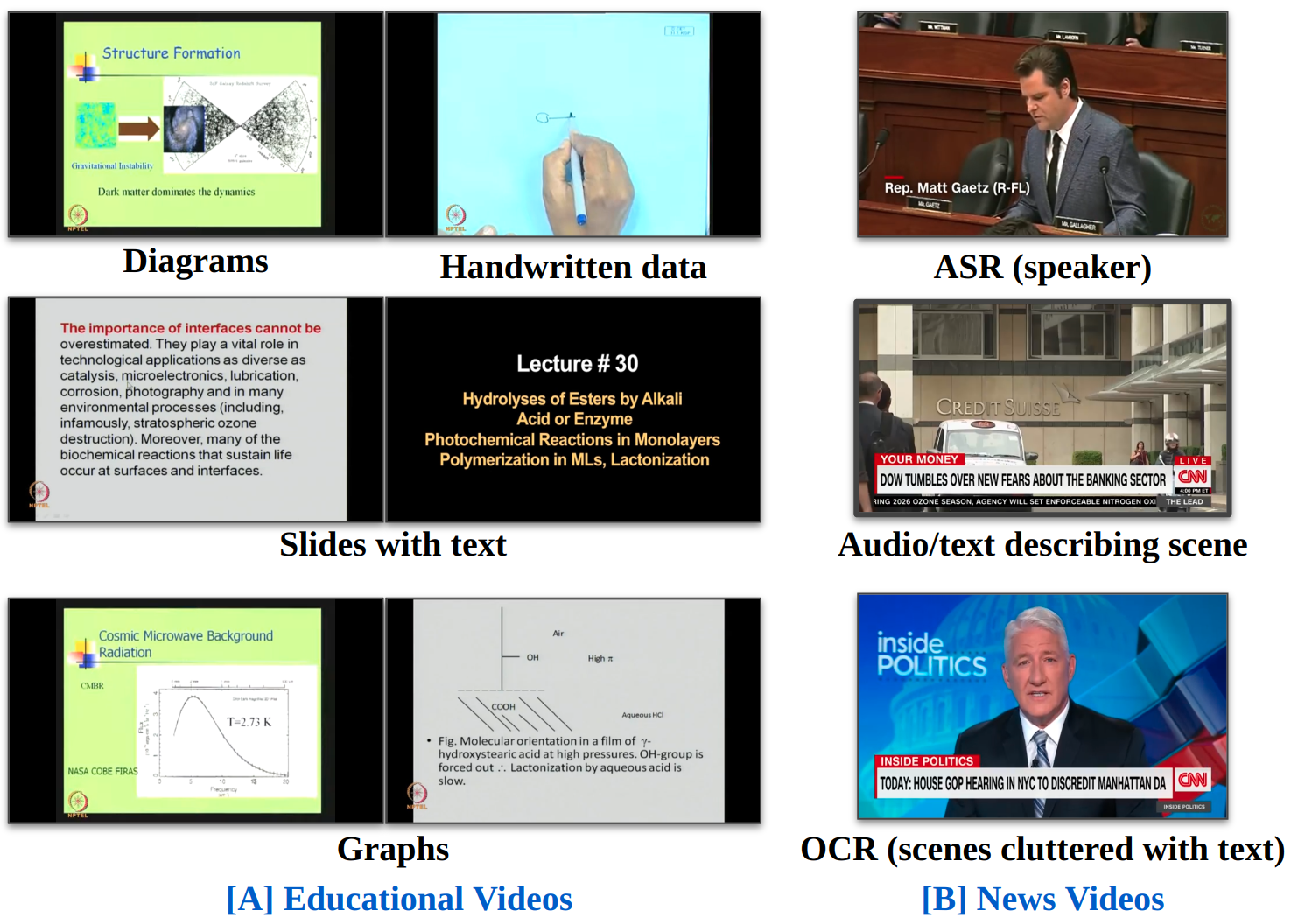}
    \caption{\textbf{An illustration showcasing the diverse data found within educational and news videos adds intrigue to these domains. Exploring such data holds promise for advancing long video understanding tasks.}
}
    \label{fig:task_intro_motivation}
\end{figure*}

\noindent
Consequently, comprehending such content practically requires extracting such live updates to be captured in text form related to the frames within the video. Optical Character Recognition (OCR) may provide useful information in such cases.
Existing video-text repositories typically consist of clips ranging from approximately 2 to 10 seconds in length, or they are trimmed to fit within this duration in extensive video datasets. However, this duration may prove ineffective for educational and news videos. Processing clips of uniform size across both domains may be challenging, as educational content often requires extended explanations for specific illustrations, spanning several minutes. Conversely, news videos deliver information swiftly, with updates occurring rapidly. Additionally, special news broadcasts may introduce diverse information with each frame, distinguishing them from the content in the analyzed clips.
ASR yields extensive textual data, but it may not be optimal for video-text pre-training due to its overwhelming nature. On the other hand, OCR may need more comprehensive information. Merging these sources could result in the loss of valuable data. Therefore, there is a need for processed information that is tailored for optimal video-text comprehension.
Recent advancements in natural language processing and computer vision have led to significant progress in developing large language models (LLMs), profoundly impacting tools within the natural language processing domain. Models like ChatGPT \cite{chatgpt}, LLaVA \cite{llava}, and others can understand and generate human-like text across diverse contexts. Leveraging LLMs to prompt and extract meaningful insights from raw datasets holds immense potential. This study delves into the extraction of additional information through carefully crafted prompts. The extracted insights offer substantial promise for future tasks, including summarization, keyword extraction, identifying key takeaways, and providing valuable metadata that can be repurposed across a broad spectrum of applications tailored to specific downstream tasks.\\
\noindent
This study aims to pinpoint the prevalent challenges associated with comprehending extended video content, particularly within educational and news contexts \Cref{fig:task_intro_motivation}. Even though captions accompany individual videos, they often need to catch up in capturing every crucial detail within lengthy videos. For instance, videos labeled with ``clustering," focusing on partition clustering, may fail to encompass the entire scope or relevance for those seeking information on subspace clustering techniques. Moreover, there are instances where students may seek particular queries, which are eventually discussed within a video rather than as its primary focus. Likewise, news videos encompass a range of topics, often accompanied by captions highlighting key events or breaking news. However, crucial information may only be captured if it coincides with the highlighted segments or breaking news alerts.
 
In such scenarios, conventional retrieval systems may not yield optimal results. Similarly, videos contain various news, captioned with highlights or breaking news; in such cases, some important information might only be caught if the breaking news or highlights.
 
Therefore, a way to understand long-form videos is needed. ASR and OCR may provide some helpful information for long-form videos.
However, the ASR transcripts are often lengthy and must be more concise to capture meaningful captions for each video clip. Conversely, OCR needs help to extract useful information from complex content, such as equations found in specific educational courses. In summary, our contributions are threefold:
\begin{enumerate}
    \item We investigate the utility of prompt templates for processing text content derived from both ASR and OCR, enabling the utilization of powerful large language models (LLMs) to generate high-quality video-text data with minimal human intervention.
    
    \item To understand the associated challenges in long-form videos, we have curated the Edu-News dataset to enhance multimodal understanding within the domains of education and news through video-language modeling.
    
    \item The analysis of baseline methods for video-text retrieval aims to comprehend the intricacies of the dataset and actively fosters interest in the research community toward this evolving direction.
\end{enumerate}

\section{Related Works}
\textbf{Video-Language Datasets.} 
Research in the domain of video-language interaction has predominantly concentrated on analyzing movies~\cite{dataset_for_movie_description_cvpr_2015,learning_visual_movie_understanding,papalampidi2021movie} and instructional content~\cite{rouditchenko2021avlnet,howto100m_iccv_2019,alayrac2016unsupervised}, notably emphasizing instructional videos related to cooking~\cite{zhou2018weakly,das2013thousand}. Despite this emphasis, a limited number of studies~\cite{hierarchial_toc_gen_edu_vid,local_recog_text_in_lecture_videos,gandhi2015topic,bulathwela2021peek,bulathwela2020vlengagement} have explored applications within the educational and news domain, which we outline below.
Specifically, certain works~\cite{hierarchial_toc_gen_edu_vid,local_recog_text_in_lecture_videos} have delved into methods for localizing and recognizing text within educational settings, particularly on blackboards. Others have examined learner engagement with educational videos~\cite{bulathwela2021peek,bulathwela2020vlengagement}, while another subset of research has focused on segmenting lecture videos~\cite{gandhi2015topic,_2023_WACV}. 
News videos frequently present a complex and multifaceted narrative, often depicting multiple events unfolding simultaneously or in quick succession. While there are some datasets tailored for understanding news content, they typically focus on specific aspects like video timeline modeling or video question answering. Recently, Chou et al.~\cite{chou2024multimodal} introduced a novel approach with long-form news videos, each encapsulating a single coherent story. These videos comprise action shots capturing real events, interviews with relevant individuals, and footage from related occurrences. However, most of these datasets necessitate manual annotation by annotators to serve specific downstream tasks.
We provide a succinct overview of existing datasets in \Cref{tab:summary}, offering insights into the scope and characteristics of available resources.


\begin{table}[htb!]
\centering
\caption{A comprehensive description of existing datasets}
\resizebox{\textwidth}{!}{%
\begin{tabular}{lcccc}
\hline
Dataset           & Content   & \# Videos & Total (hrs) & Annotations \& Others                                 \\ \hline
Reuters ViL News & News      & 1970                         & 50          & Subtitle, Caption, Story Keyword                     \\
NewsVQA          & News      & 2730                         & 156         & QA Pairs                                               \\
YT-News Timeline & News      & 300K                         & -           & Timelines                                              \\
NewsVideo QA     & News      & 3083                         & -           & QA, OCR                                                \\ \hline
LectureVideoDB   & Education & -                            & -           & \begin{tabular}[c]{@{}c@{}}Word bounding boxes and\\ corresponding text ground truths\\ for 5000 frames\end{tabular} \\
AVL Lectures     & Education & 2350                         & 2200        & OCR \& Transcripts                                    \\
\textbf{Ours}    & Education+News & 5146                    & 1500+       & \begin{tabular}[c]{@{}c@{}}OCR \& ASR Transcripts,\\ Caption using Prompts\end{tabular} \\ \hline

\end{tabular}
}
\label{tab:summary}
\end{table}

\noindent
\textbf{Prompt Engineering.}
Prompt engineering~\cite{liu2023pretrain,white2023prompt} represents a cutting-edge methodology in natural language processing (NLP), revolutionizing the utilization of large language models (LLMs) by enhancing efficiency and cost-effectiveness. This groundbreaking paradigm stems from the evolution of LLMs, which have revolutionized our natural language comprehension. Their adeptness at discerning linguistic patterns and structures has proved indispensable across language-centric endeavors. Nevertheless, fine-tuning these pre-existing models can be resource-intensive and time-consuming, necessitating substantial annotated data and computational power.
In response to this challenge, researchers have embraced prompts to guide model refinement~\cite{white2023prompt,lester2021power}. Prompt learning, an innovative framework within NLP, empowers language models to engage in few-shot or even zero-shot learning, seamlessly adapting to novel contexts with minimal labeled data. This methodology, deeply rooted in language modeling, entails directly modeling the probability of text. The meticulous design of prompts tailored to specific downstream tasks is at the core of prompt engineering, effectively steering pre-trained models toward accomplishing desired objectives. Recently, downstream tasks related to images and videos have begun leveraging prompting technology, either through manually crafted prompts or automatically generated prompts, to yield improved results in multimodal learning and enhance zero-shot performance~\cite{wasim2023vitaclip,ju2022prompting,chen2023groundingprompter}.
\section{Edu-News: A prompt based Multimodal Dataset}

\subsection{Data Curation}
In the realm of video understanding, predominant research endeavors are concentrated on instructional videos, while educational and news content remains comparatively underexplored despite presenting distinct challenges. Educational videos are characterized by essential elements such as blackboards, PowerPoint presentations, and instructor narration. Conversely, news videos encompass a diverse array of critical information, including breaking news, updates, live coverage, and other significant events. 
\noindent

\begin{figure*} 
    \centering
    \includegraphics[width=0.85\linewidth]{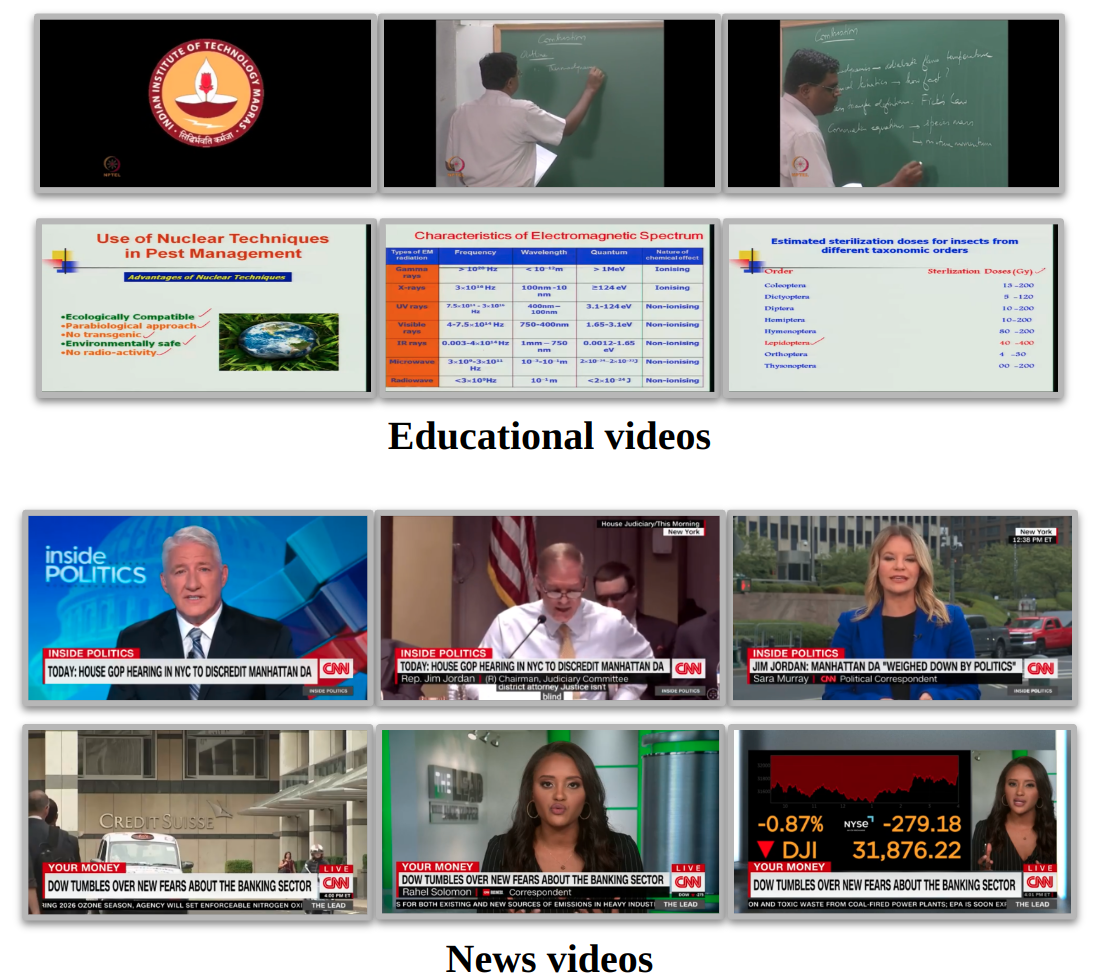}
    \caption{\textbf{Edu-News: Dataset for video understanding on educational and news videos.}
}
    \label{fig:task_intro}
\end{figure*}
\begin{figure*} 
    \centering
    \includegraphics[width=1\linewidth]{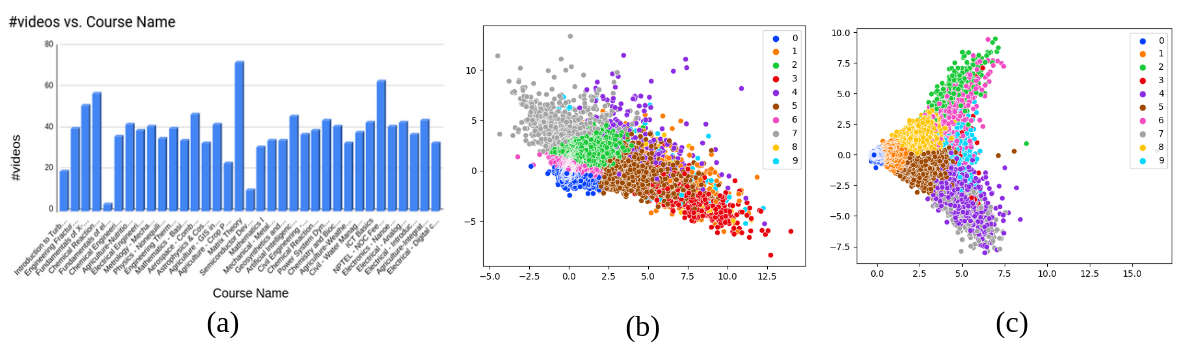}
    \caption{Insights into the distribution of data investigated in this study are provided. In (a), the distribution of topics in NPTEL videos is illustrated, demonstrating a uniform spread across topics. Notably, the experiments outlined in this work can extend to encompass a broader range of topics and diverse lecture video content.  In (b) and (c), we depict the distributions of word embeddings of OCR tokens and transcripts, respectively, in both educational and news videos. We employ clustering techniques with k=10 clusters to visualize the diverse array of video content types present in the dataset.
}
    \label{fig:stat}
\end{figure*}
\begin{figure*} 
    \centering
    \includegraphics[width=1\linewidth]{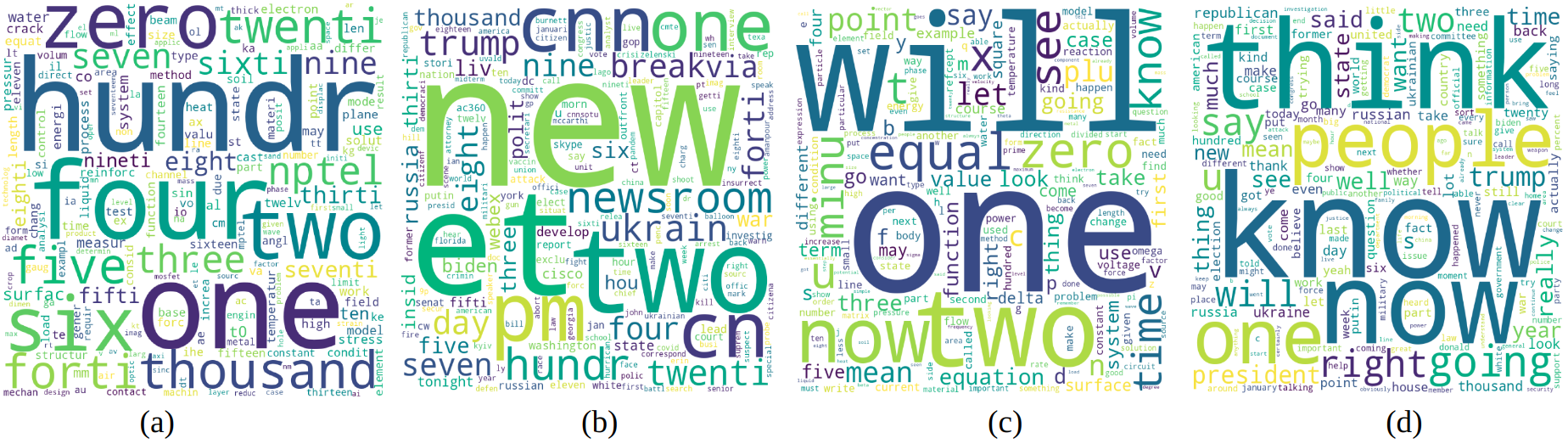}
    \caption{Word cloud representation of video content in the dataset. (a) Illustrates the word distribution within OCR tokens extracted from (a) NPTEL videos and (b)   news videos. Displays the word distribution within transcripts of (c) NPTEL videos and (d)  news videos.
}
    \label{fig:word_clouds}
\end{figure*}

\textbf{Collection Strategies.}
SWAYAM-NPTEL stands as a flagship initiative by the MHRD, aimed at delivering top-tier education nationwide to all learners. We have meticulously selected a range of diverse courses covering engineering, science, and technology, leveraging the utilization of blackboards and PowerPoint presentations to curate multimodal educational content, enhancing video comprehension for learners. 
Moreover, we collect news videos from multiple channels, mainly categorized into US and World news. They cover various topics, including politics, current affairs, entertainment, and sports. We curate individual YouTube playlists for both news and educational videos.\\
\noindent
\textbf{Statistics.}
In our analysis of lecture videos, we compile a selection of 35 courses from NPTEL, encompassing a total of 1345 videos, with an average of 38 videos per course. The distribution of courses is illustrated in \Cref{fig:stat}, which presents a bar chart indicating the number of videos across all 35 courses from NPTEL. Similarly, for news videos, we collect approximately 1000 hours of content.\\
\noindent
\textbf{Features.} 
Lecture videos undergo content changes less frequently due to the speaker's detailed explanations of concepts or PowerPoint presentations. In contrast, news videos exhibit rapid content turnover, with significant fluctuations in textual content at the frame level. It's essential to pinpoint specific usable and distinct frames to extract OCR-like features. The strategy for selecting frames differs between education and news videos. For long-form education videos averaging around 38 minutes, we sample one frame per minute, whereas for news videos, we sample one frame per 10 seconds. We employ the EasyOCR library, an open-source tool~\cite{easyocr}, to extract OCR content from the chosen frames. We employ the EasyOCR library, an open-source tool~\cite{easyocr}, to extract OCR content from the chosen frames. Furthermore, we utilize transcripts, subtitles, and closed captions to understand the video content comprehensively. We extract these speech-to-text components by employing the Silero~\cite{silero_models} model provided in PyTorch. These transcripts serve a dual purpose: they contribute to the retrieval process and form the foundation for our prompting methodology. In \Cref{fig:word_clouds}, we show the word cloud of OCR tokens and Transcripts in the dataset. \\

\begin{figure*} 
    \centering
    \includegraphics[width=1\linewidth]{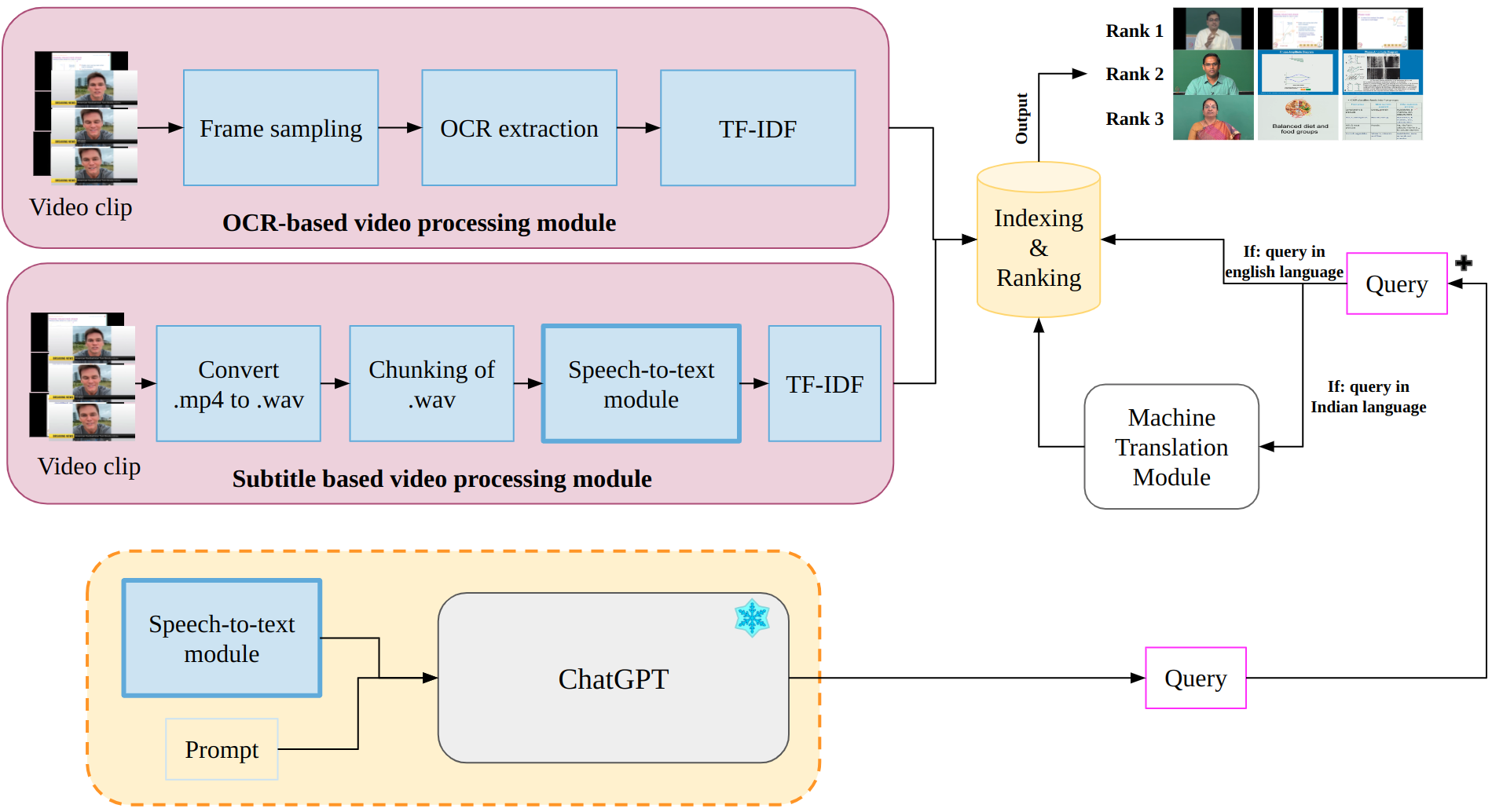}
    \caption{Pipeline Overview: Initial processing involves leveraging multimodal cues such as OCR tokens and transcripts to analyze long-range videos. Subsequently, retrieval is performed using either captions generated by ChatGPT or alternative queries. Additionally, our system accommodates queries in Indian languages. 
    }
    \label{fig:pipeline}
\end{figure*}

\subsection{Designing prompt templates}
\begin{figure}[htb!]
    \centering
        \includegraphics[width=\textwidth]{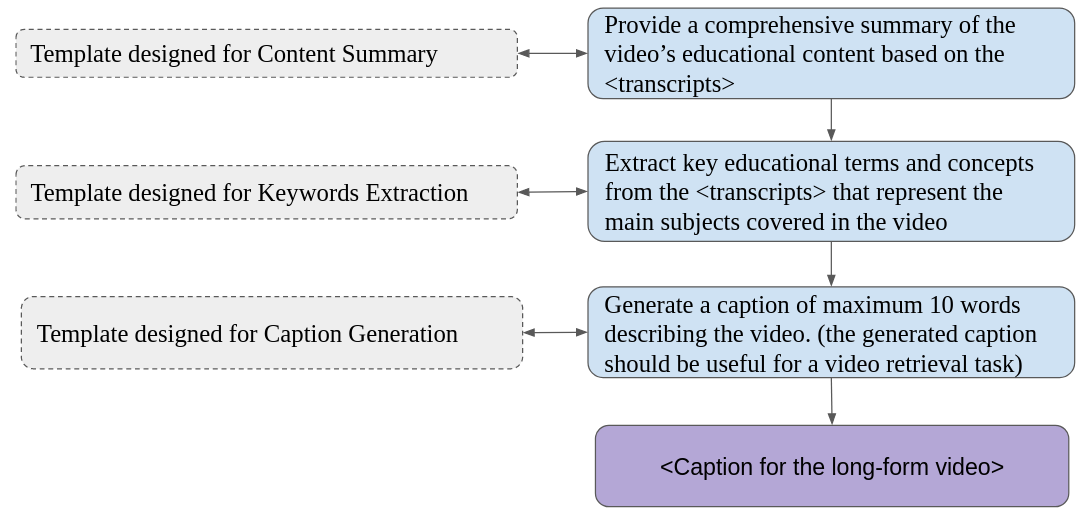}
        \caption{A collection of prompt templates utilized to generate caption for each video from the transcript generated by ASR outputs extracted from each long-form educational video.}
        \label{fig:edu_prompt}
\end{figure}

\begin{figure}[htb!] 
        \centering
        \includegraphics[width=\textwidth]{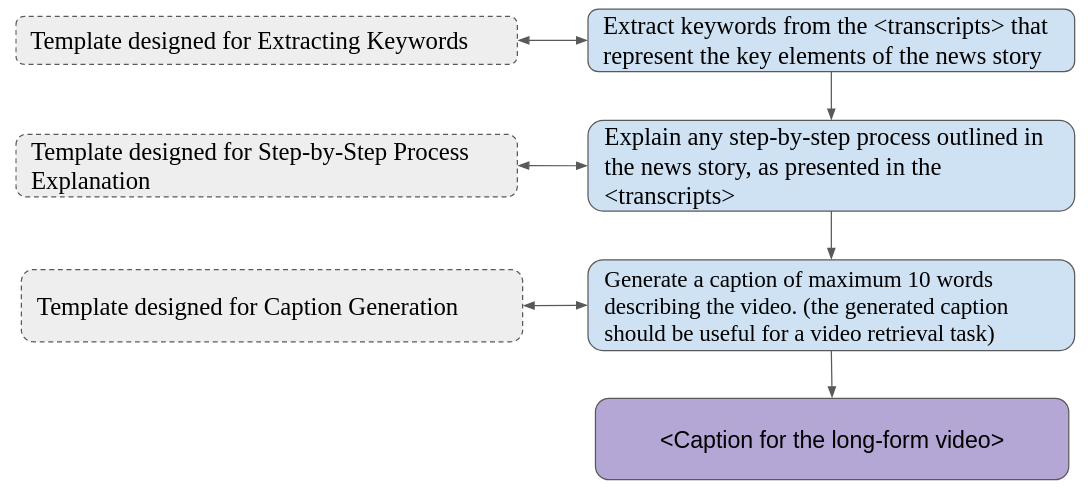}
        \caption{A collection of prompt templates utilized to generate caption for each video from the transcript generated by ASR outputs extracted from each long-form news videos.}
        \label{fig:news_prompt}
\end{figure}

As computing capabilities advance, contemporary research increasingly relies on leveraging large language models trained on extensive datasets. One method involves prompting these models with initial text inputs, such as with ChatGPT, Bard, LLaVa, and others, enabling them to produce coherent and contextually relevant content. This approach resembles initiating a creative dialogue with the model, where the input prompt is a guiding seed for generating responses. The resulting data demonstrates the model's proficiency and ability to emulate diverse styles, foster creativity, and offer valuable insights.\\
\noindent
In our experiments, we employed large language model: ChatGPT 3.5, developed by OpenAI. This model havs been trained on diverse and extensive textual datasets, enabling them to comprehend and generate human-like text across various domains and contexts. It can perform multiple tasks, including generating text, translating languages, and providing informative responses to questions. Additionally, It can follow instructions, complete requests thoughtfully, and answer questions comprehensively, even when faced with open-ended and challenging inquiries. Moreover, such models are capable of generating text in different creative formats, such as code, scripts, letters, poems, and more.
In this work, we leverage the potential of prompting techniques for generating content in the context of long-range educational and news video understanding.
However, these two types of videos vary significantly in content, necessitating the inclusion of tailored prompts to extract the most suitable captions for each video.
We establish a distinct set of prompts corresponding to the respective video domains, individualized prompts for educational content, and separate prompts for news content. Depending on the input provided to these models, we further finetune the prompts to capture the necessary essence, ensuring alignment with the unique characteristics of the data at hand.
\Cref{fig:edu_prompt,fig:news_prompt} display the sequential series of prompt templates employed to extract captions from the transcripts generated through ASR for the entire lengthy video.

\section{Experiments}

\subsection{Evaluation setups}
We assess our dataset for long-form video retrieval performance with natural language queries. The objective is to identify relevant videos from a vast collection based on textual descriptions. We gauge the effectiveness of our learned embedding using standard recall metrics such as R@1, R@5, and R@10.
\subsection{Comparison with state-of-the-art}
\textbf{Term Frequency-Inverse Document Frequency (TF-IDF).} It is a method employed in information retrieval and text mining to assess a document's relevance to a query. It involves calculating the Term Frequency (TF), which measures term frequency within a document, and the Inverse Document Frequency (IDF), which gauges term rarity across the document collection. TF-IDF is derived from these components and is utilized to identify documents most relevant to a given query term. This technique finds widespread application in search engines and text analysis systems.\\
\noindent
\textbf{Dense Passage Retriever (DPR)~\cite{dpr}.} The method employs a dense encoder $E_p$, which transforms any textual passage into $d-$dimensional real-valued vectors and constructs an index for all $M$ passages utilized in retrieval. During inference, DPR utilizes a distinct encoder $E_q$ to map the input question to a d-dimensional vector, retrieving k passages whose vectors closely align with the question vector.\\
\noindent
\textbf{Singularity~\cite{singularity_acl_2023}.}
The method explores single-frame training methodologies for tasks involving video and language. The study demonstrates state-of-the-art performance in text-to-video retrieval by leveraging substantial pre-training data and employing an effective multi-frame ensemble strategy during inference.

\subsection{Video retrieval system}
We present a web application for a versatile video retrieval system capable of handling diverse tasks across news and educational domains. Our system accommodates a range of search queries, from simple single-keyword inquiries to more complex searches involving multiple keywords or complete sentences (such as those generated by ChatGPT). The search process is refined by eliminating stopwords and tokenizing sentences, laying the groundwork for our retrieval mechanism.
Our approach encompasses two distinct paths, depicted in \Cref{fig:pipeline}: one leveraging OCR tokens and the other utilizing transcripts or closed captions. In the OCR path, we preprocess OCR tokens obtained using EasyOCR by removing stopwords, punctuation, and irrelevant details before storing them in a database. These tokens serve as the foundation for our video search engine's functionality. Concurrently, the transcript path involves extracting transcripts or closed captions using a Speech-to-text model, followed by preprocessing to ensure consistency and relevance akin to OCR tokens.
For retrieval, OCR tokens and transcripts are separately employed within a TF-IDF-based algorithm. The system provides the flexibility to choose between OCR token-based or speech-to-text-based retrieval. Furthermore, it supports queries in multiple Indian languages, translating them into English before initiating video retrieval operations.

\subsection{Quantitative results}
Predominantly existing text-video retrieval systems and datasets work on small-duration videos (approx. 10 sec) and have considered only visual content in the videos to generate captions/queries. Conversely, in the proposed dataset, the news and educational videos span up to approximately 45 minutes. These videos contain a wide range of content majorly originating from textual (OCR) and audio (transcripts), making it challenging to apply existing SOTA methods. Hence, evaluating such methods could raise concerns about their reliability. It is very important to note that creating precise ground truth data: ``video-caption" pairs for lengthy videos is an expensive task. However, we have conducted a comparison of the Dense Passage Retrieval (DPR) text-only model \cite{dpr} and SINGULARITY \cite{singularity_acl_2023}, a vision-language multimodal method, and its variants against the proposed method TF-IDF-based on this dataset.

\begin{table}[h]
    \centering
    \footnotesize 
    \caption{ Quantitative results for a randomly sampled test set of 60 video-caption pairs. Captions are generated by Humans and ChatGPT}
    \resizebox{\columnwidth}{!}{%
    \begin{tabular}{@{}lcccccrrccrr@{}}
        \toprule
        & & \multicolumn{3}{c}{\textbf{Human-caption}} & \multicolumn{3}{c}{\textbf{ChatGPT-caption}} \\
        \cmidrule(lr){3-5} \cmidrule(lr){6-8}
        \textbf{Method} & \textbf{Features} & \textbf{R@1} & \textbf{R@5} & \textbf{R@10} & \textbf{R@1} & \textbf{R@5} & \textbf{R@10} \\
        \midrule
        Random & - & 01.67 & 11.67 & 20.00 & 01.67 & 11.67 & 20.00 \\
        DPR & OCR & 01.67 & 33.33 & 20.00 & 03.33 & 20.00 & 30.00 \\
        DPR & transcripts & 00.00 & 06.67 & 18.33 & 01.67 & 06.67 & 13.33 \\
        SINGULARITY-5m & visual & 16.66 & 33.33 & 51.66 & 16.66 & 43.33 & 53.33 \\
        SINGULARITY-17m & visual & 20.00 & 40.00 & 53.33 & 18.33 & 43.33 & 56.66 \\
        SINGULARITY-5m-tmp & visual & 13.33 & 43.33 & 50.00 & 21.66 & 40.00 & 58.33 \\
        SINGULARITY-17m-tmp & visual & 15.00 & 33.33 & 46.66 & 16.66 & 38.33 & 56.66 \\ \midrule
        \textbf{TF-IDF-based (Ours)} & transcripts & {38.33} & {46.67} & {50.0} & \textbf{40.00} & {48.33} & {58.33} \\
        \textbf{TF-IDF-based (Ours)} & OCR & \textbf{40.00} & \textbf{51.67} & \textbf{60.00} & 36.67 & \textbf{55.00} & \textbf{66.67} \\
        \textbf{TF-IDF-based (Ours)} & transcripts+OCR & \textbf{40.00} & {50.00} & {56.67} & 36.67 & {53.33} & {61.67}\\
        \bottomrule
    \end{tabular}%
    }
    \label{tab:quant_cap}
\end{table}

We do this by annotating a \textbf{test set} of 60 video-caption pairs. We generate captions manually and by ChatGPT. To generate captions using ChatGPT, we prompt it to generate captions using the transcripts of the videos. From \Cref{tab:quant_cap}, it can be seen that both DPR and SINGULARITY underperform compared to the proposed method. i) DPR can consider a maximum input size (512) which is insufficient for the context of long-range videos. Hence to evaluate, we randomly sample words from the context and truncate it to the max length. ii) SINGULARITY is a multimodal method. It considers 4 frames at the time of testing to retrieve a video. This results in sparse data usage similar to DPR which is not sufficient to retrieve long-range videos such as educational videos and news videos. On the contrary, a simple non-learning-based TF-IDF-based method can do efficient retrieval by considering information present from multiple frames (both textual and audio in separate and combined forms) throughout the long-range video. It is also important to note that DPR and SINGULARITY are tested for zero-shot performance since, we do not have training data associated for the proposed dataset.

\begin{figure*} 
    \centering
    \includegraphics[width=.85\linewidth]{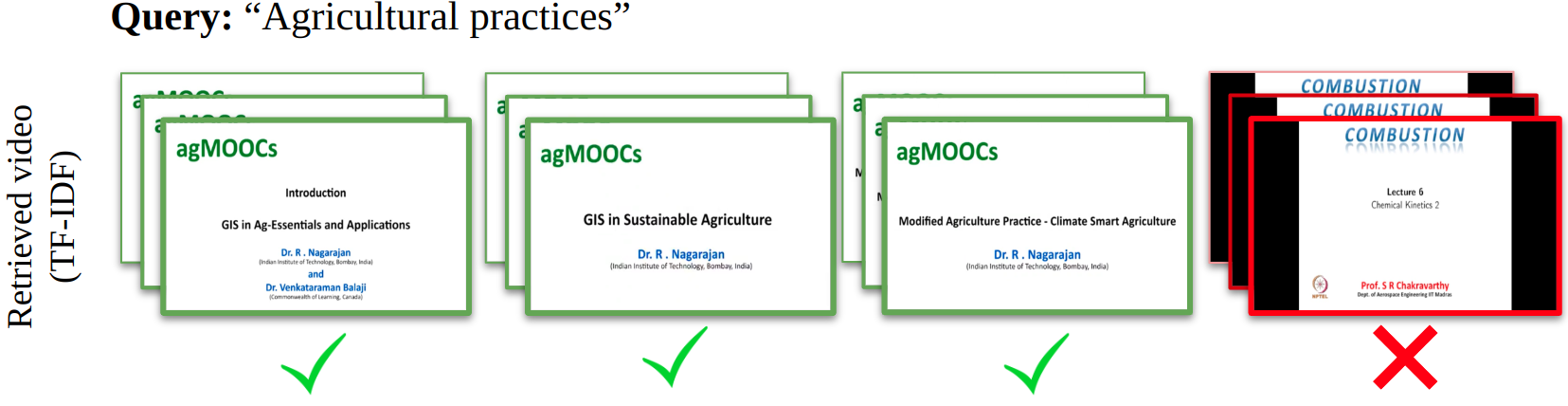}
    \caption{Results of the proposed method when input is a keyword like ``Agricultural practices". It can be seen that the method retrieves relevant videos (shown in green).}
    \label{fig:qual_res_2_keyword}
\end{figure*}

As seen in \Cref{tab:quant_cap}, the utilization of DPR (Dense Passage Retrieval) in this context is multifaceted, as it retrieves outputs based on both Optical Character Recognition (OCR) and transcripts, as exemplified in the table. However, a notable challenge lies in the fact that conventional DPR models are typically trained on text corpora with significantly smaller contextual scopes, resulting in cleaner data compared to the complex nature of OCR and transcripts. Singularity, on the other hand, operates as a video retrieval model, primarily trained on 10-second video clips for retrieval purposes, with a focus on the actions depicted within these clips. Yet, when confronted with longer videos, a workaround involves clipping the initial 10 seconds for retrieval, which poses difficulties due to the disparity between the short clips and the comprehensive captions derived from the entire length of the video. In contrast, TF-IDF (Term Frequency-Inverse Document Frequency) exhibits superior performance, as it encompasses content from the entirety of the video rather than solely focusing on brief segments. Collectively, these challenges underscore the complexity inherent in the proposed dataset, necessitating innovative solutions to address the inherent limitations of existing models and techniques.

\begin{figure*} 
    \centering
    \includegraphics[width=.85\linewidth]{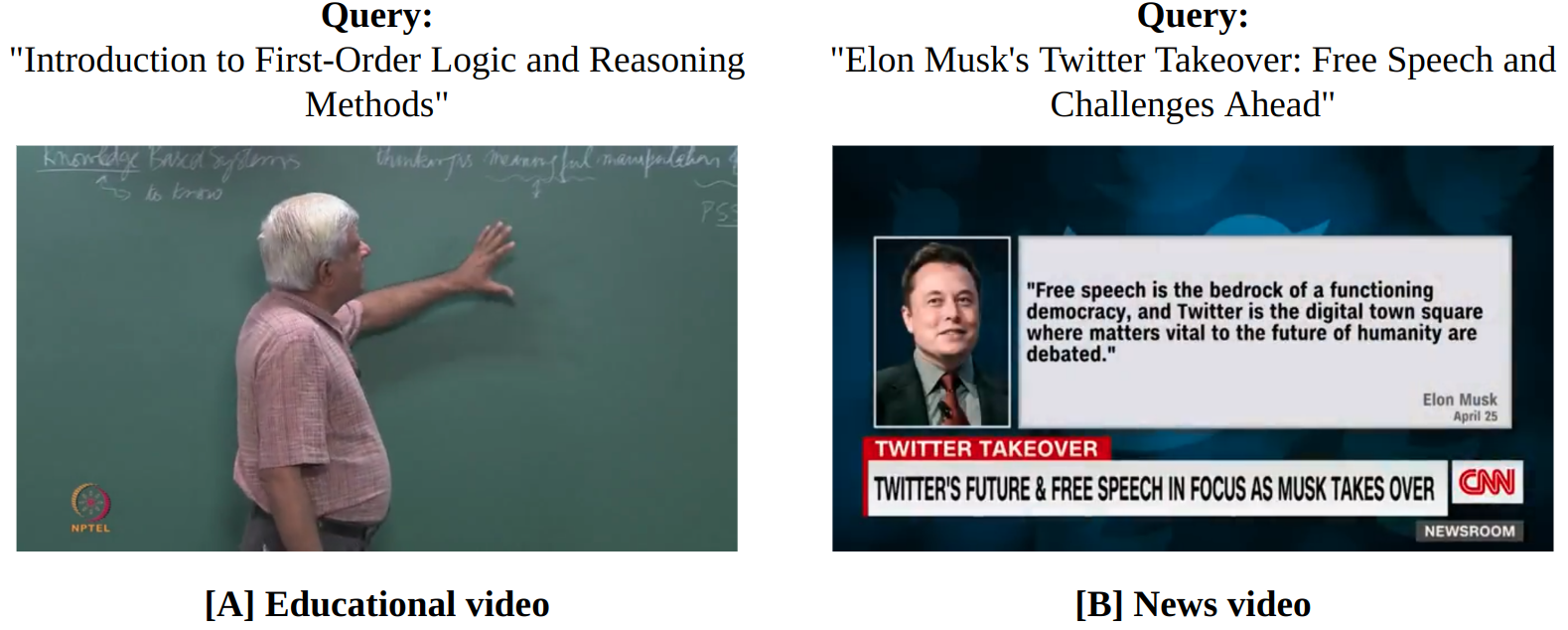}
    \caption{Results of the proposed method when input is a caption generated by ChatGPT. It can be seen that a simple method can retrieve the desired output.}
    \label{fig:qual_res_chatgpt_prompt_full_data_tfidf}
\end{figure*}

\begin{figure*} 
    \centering
    \includegraphics[width=0.85\linewidth]{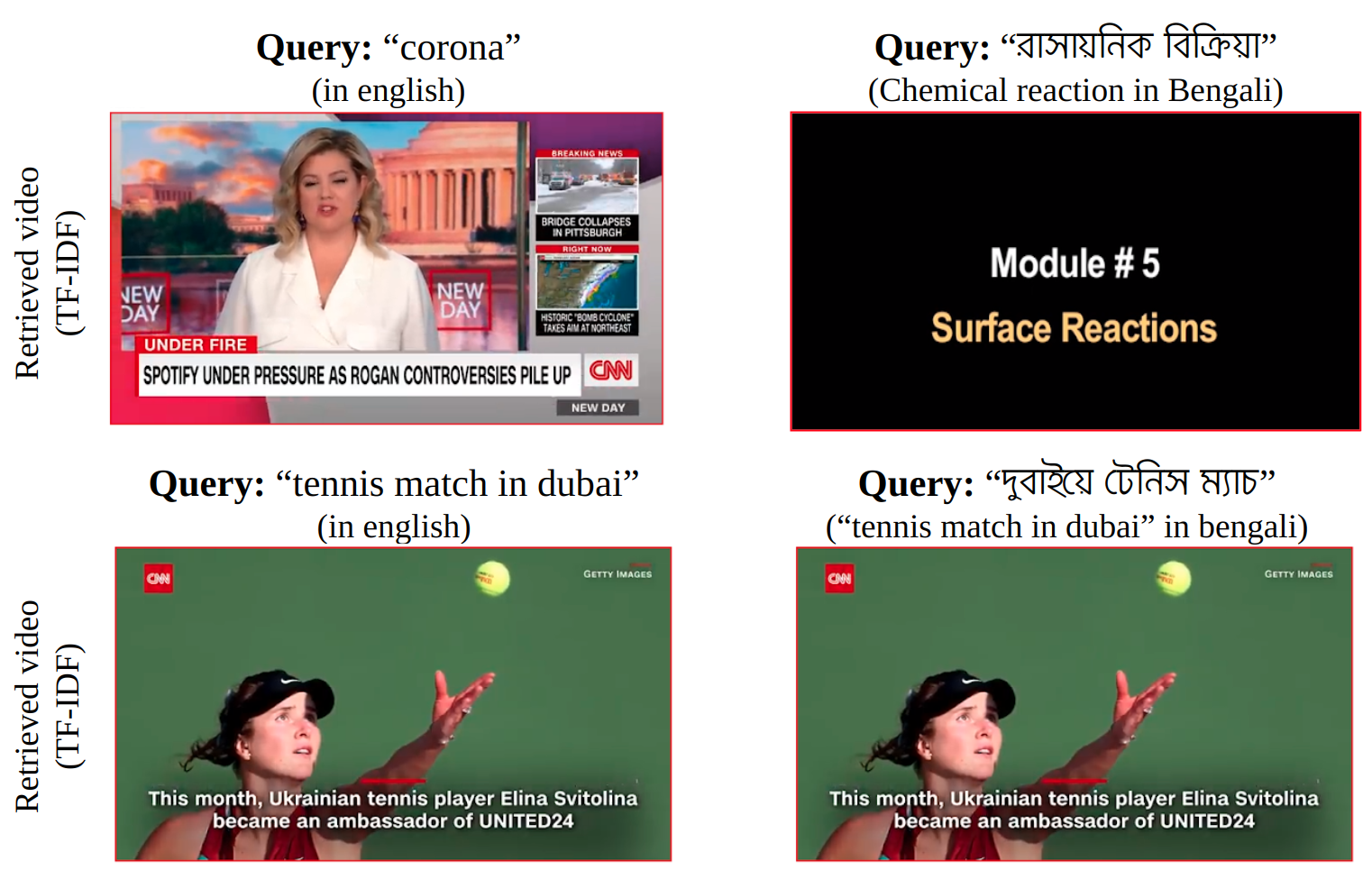}
    \caption{Qualitative results for queries in Indian languages are showcased. In the first example, the query term "corona" is provided in English. In the second example, we input chemical reactions in Bengali, which the system translates into English for retrieval. In the third and fourth examples, "tennis match in Dubai" is presented in both English and Bengali. }
    \label{fig:ocr_qual_tennis_eng_bangla}
\end{figure*}





\subsection{Qualitative results}

In \Cref{fig:qual_res_chatgpt_prompt_full_data_tfidf}, we illustrate the results obtained from the TF-IDF-based retrieval system. The input query for this retrieval process consists of captions generated by ChatGPT, derived from the transcripts or closed captions of a diverse range of content, including both news and educational videos. This approach aims to leverage the richness of information embedded within the video content to enhance the accuracy and relevance of the retrieved outputs.

The qualitative results for queries in Indian languages are demonstrated in \Cref{fig:ocr_qual_tennis_eng_bangla}. In the first instance, the query term "corona" is presented in English, showcasing the system's capability to handle queries in the English language. Moving forward, the second example illustrates the system's proficiency in translating Bengali queries regarding chemical reactions into English for effective retrieval. Expanding further, the third and fourth examples depict queries about "tennis match in Dubai," showcased in both English and Bengali, highlighting the system's versatility in understanding and processing multilingual queries. This diversity underscores the system's adaptability to cater to users across linguistic preferences and aids in fostering inclusive information retrieval experiences.



\section{Conclusion}
With the rapid advancement of digital media content, there has been a notable surge in the creation of long-form videos online, particularly in education and news. Consequently, comprehending these lengthy videos has garnered increased attention within the research community. We introduce the Edu-news dataset by Utilizing various informational cues, including OCR tokens and transcripts. This dataset is a foundation for exploring potential avenues to augment long-form video comprehension through prompt engineering. Our baseline findings reveal limitations when applied to long-form video datasets. For instance, existing text-based query systems struggle to process large text data efficiently. Additionally, video text retrieval systems exhibit poor quality, as they only consider the initial few seconds of video, resulting in subpar retrieval outputs.

\bibliographystyle{plain}
\bibliography{ref} 
\end{document}